\documentclass[letterpaper]{article} 
\usepackage{aaai2026}
\usepackage{times}  
\usepackage{helvet}  
\usepackage{courier}  
\usepackage[hyphens]{url}  
\usepackage{graphicx} 
\usepackage{natbib}  
\usepackage{caption} 
\usepackage{algorithm}
\usepackage{algorithmic}
\usepackage{amsmath}

\usepackage{newfloat}
\usepackage{listings}
\DeclareCaptionStyle{ruled}{labelfont=normalfont,labelsep=colon,strut=off} 
\floatstyle{ruled}
\newfloat{listing}{tb}{lst}{}
\floatname{listing}{Listing}
\title{Polaris : Multi Agentic System for Conversational Enterprise Analytics}

\begin{document}

\author{
Varuni H K,
Soham Sarkar,
Jay Kumar,
Goutham Krishnan,
Tanvi Johari,
Avinash Bharadwaj,
Santosh Hegde
}

\affiliations{
Couchbase Inc.\\
India\\
\{varuni.hk, soham.sarkar, jay.kumar, goutham.krishnan, tanvi.johari, santosh.hegde, avinash.bharadwaj\}@couchbase.com
}

\maketitle

\begin{abstract}
In today’s fast-paced environment, the ability to swiftly access, understand, and act on data is no longer optional, it is essential. Yet most organizations remain data-rich but insight-poor, constrained by the complexity of querying, interpreting, and explaining enterprise-scale information. We present Polaris, a supervisor-led multi-agent framework for conversational enterprise analytics that bridges this gap. Polaris introduces Dynamic Task Coordination (DTC), a decision-theoretic orchestration layer that models agent–task assignment as adaptive bipartite matching, enabling real-time coordination, recovery, and optimization across specialized agents for querying, visualization, and reasoning. By coupling DTC with reason-first, ReAct-style agents, Polaris transforms natural language queries into coherent analytical workflows that not only retrieve and visualize data but also explain the underlying “why.” Evaluation on structured enterprise datasets demonstrates high semantic fidelity and answer relevancy, underscoring the potential of multi-agent orchestration to deliver trustworthy, end-to-end business intelligence at scale.
\end{abstract}


\section{Introduction}

The proliferation of large-scale enterprise data repositories has created unprecedented opportunities for data-driven decision making, yet the complexity of extracting actionable insights from heterogeneous data sources remains a fundamental bottleneck in organizational intelligence systems. Traditional approaches to enterprise data analytics rely heavily on domain expertise in query languages, schema understanding, and statistical interpretation, creating significant barriers to widespread adoption and limiting the democratization of data-driven insights across organizational hierarchies.

Large language models (LLMs) enable natural language interaction, but most systems rely on single agents that struggle with compositional reasoning, multi-step coordination, and session-level coherence \cite{wang2023survey,xi2023rise}. These limits are acute in enterprise settings that require query generation, statistical analysis, visualization, and narrative explanation.

Multi-Agent Systems (MAS) decompose complex tasks into coordinated components and can outperform single agents in tool use and long-horizon planning \cite{stone2000multiagent,tampuu2017multiagent,wu2023autogen,hong2023metagpt,ReAct}. Yet MAS for enterprise analytics remains relatively underexplored, especially adaptive coordination that optimizes task allocation and enables intelligent error recovery in conversational workflows.

We propose a supervisor-led MAS framework with Dynamic Task Coordination (DTC): 
Dynamic Task Coordination (DTC), a decision-theoretic, online orchestration layer for conversational data analytics. 
DTC models coordination as adaptive bipartite matching on $G = (A \cup T, E)$, where each agent $a \in A$ has a capability vector $\boldsymbol{\theta}_a$ and each task $t \in T$ has a descriptor $\boldsymbol{\tau}_t$. Edge utilities combine capability fit, and empirical reliability: $U(a,t \mid s) = \alpha \, \text{sim}(\boldsymbol{\theta}_a, \boldsymbol{\tau}_t) + \beta \, q(a,t) - \gamma \, c(a,t)$, given state $s$. We formalize matching as:
\begin{equation}
M^* = \arg\max_{M \in \mathcal{M}} \sum_{(a,t) \in M} U(a,t \mid s)
\label{eq:dtc-matching}
\end{equation}
Here, $M^*$ denotes the optimal feasible matching (the selected set of agent--task pairs), and $\arg\max$ indicates choosing, among all feasible $M \in \mathcal{M}$, the one that maximizes the total utility. This decision-theoretic formulation maximizes capability fit and empirical reliability while penalizing cost, yielding principled allocations under constraints. $\mathcal{M}$ denotes feasible matchings under precedence, capacity, and tool/data-availability constraints. Execution feedback (diagnostic traces, errors, latencies) updates $q(a,t)$ and the belief over $\boldsymbol{\theta}_a$, enabling uncertainty-aware re-planning, preemption, and rollback. This yields real-time constraint satisfaction, adaptive load balancing, and capability-aware assignment across heterogeneous agents.


Our framework seamlessly integrates structured data analysis, dynamic visualization generation, and causal reasoning within a unified conversational interface. This multi-modal approach addresses the critical gap between data presentation and explanatory understanding that limits the effectiveness of existing business intelligence systems.

\section{Proposed Work}
At its core, Polaris employs a network of specialized agents coordinated through our Dynamic Task Coordination (DTC) framework. Concretely, the supervisor solves the matching problem in Eq.~\ref{eq:dtc-matching} as user goals, tool latencies, and data availability evolve. This centralized coordination maintains persistent state across sessions, orchestrates sub-tasks via capability-aware matching, and enforces semantic alignment between multi-modal outputs (e.g., ensuring visualizations, query results, and narratives remain coherent). 
\begin{figure*}[t]
    \centering
    \includegraphics[width=0.9\linewidth,height=0.35\textheight]{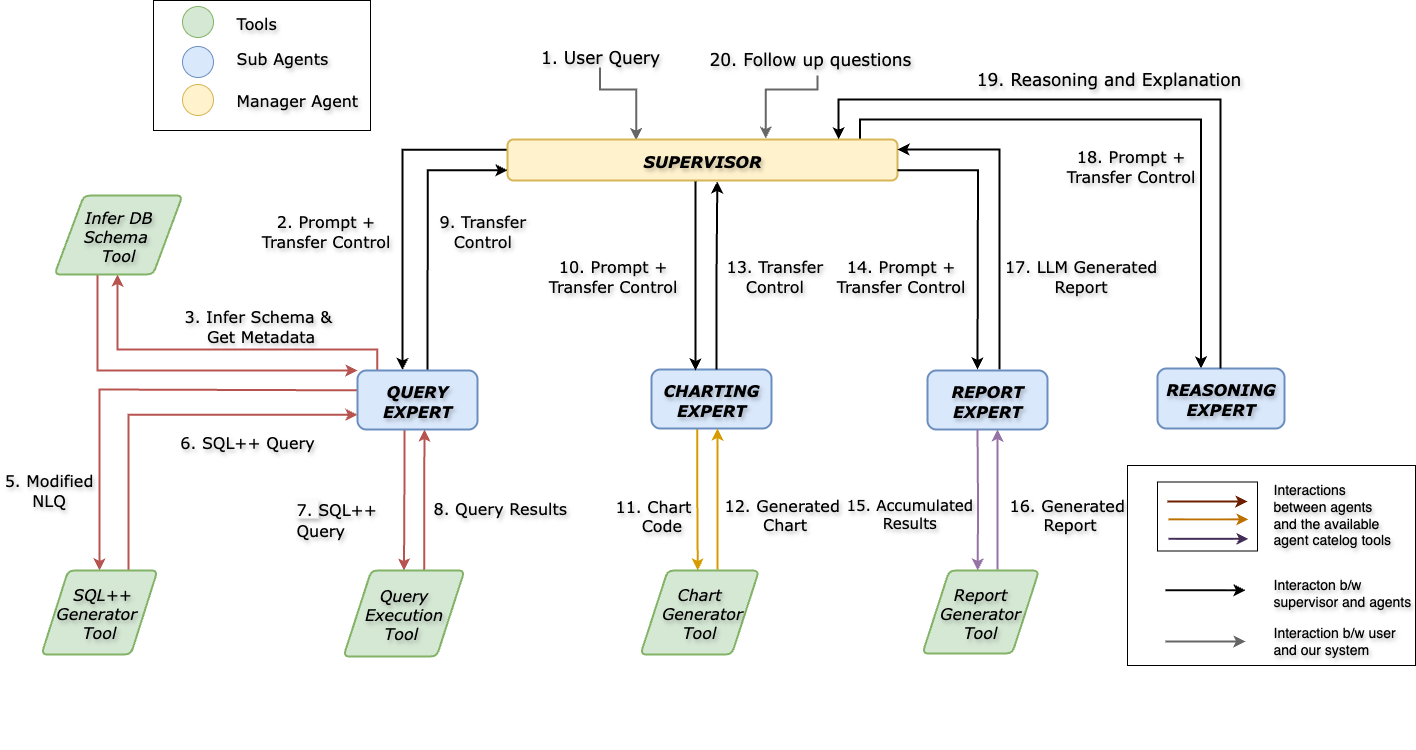}
    \caption{Architecture and Decision Flow for Polaris}
    \label{fig:placeholder}
\end{figure*}

All agents in Polaris adopt a reason-first approach within the DTC coordination paradigm, where contextual reasoning precedes tool selection and execution, following a ReAct (Reasoning + Action) framework \cite{ReAct,shen2023hugginggpt,schick2023toolformer}. By interleaving chain-of-thought reasoning with concrete actions in adaptive thought–action–observation loops, agents can intelligently plan multi-step workflows, dynamically adapt to intermediate outcomes and system constraints, and produce coherent, context-aware insights while maintaining optimal resource utilization through DTC optimization.

\subsection{Orchestration and Understanding: Supervisor Agent}

The Supervisor Agent is the main controller of the system and implements DTC end-to-end: it parses intent, maintains context, and routes tasks by approximately solving Eq.~\ref{eq:dtc-matching} under precedence and capacity constraints \cite{kuhn1955hungarian,karp1990online}. It decomposes ambiguous requests into concerete workflows (e.g., causal analysis, trend comparisons, multi-dimensional reporting). Real-time diagnostics update $q(a,t)$ and constraints, enabling re-matching, rollback, or alternative tool selection. Upon schema mismatches or missing fields, the Supervisor triggers recovery protocols, proposes semantically equivalent alternatives, and preserves conversational continuity.

\subsection{Data Retrieval and Transformation}
\vspace{-4pt}
The Query Expert addresses one of the hardest gaps in conversational data intelligence: translating ambiguous natural language into precise, executable queries in SQL++ \cite{sql++}. Its operation begins with schema inference and annotation, where it not only identifies fields and datatypes through the SQL++ INFER command but also grounds them in semantic annotations. This disambiguation is crucial, what looks like a generic field such as "amount" can be resolved into "total sales amount" when the context demands it, ensuring queries are aligned with enterprise-specific semantics rather than surface-level matches. Next, through input canonicalization, vague user prompts are expanded into explicit task statements, enabling the system to bridge the mismatch between colloquial phrasing (“sales last month”) and the structured detail required for query generation. Once canonicalized, the request undergoes NL-to-SQL++ translation, where reasoning steps are intertwined with tool calls rather than separated, following a reason-first ReAct approach \cite{ReAct}. Finally, the Query Expert performs data quality checks and adaptive recovery: it detects null-heavy columns, schema mismatches, or overlarge results, and autonomously reformulates queries. In cases where the output would exceed LLM context limits, the agent shifts to adaptive aggregation, summarizing distributions without sacrificing fidelity to trends and anomalies. The contribution of this design is subtle but significant: by combining semantic annotation, canonicalization, and reasoning-driven aggregation, the Query Expert ensures that data retrieval is not just syntactically correct but contextually faithful and resilient to the imperfections of real-world enterprise data.

\subsection{Visualization and Insights: Charting Expert}
The Charting Expert is responsible for transforming structured query outputs into visual representations that facilitate data interpretation. Unlike conventional systems that rely on predefined templates, it uses a combination of rule-based heuristics and the ReAct framework to determine the most suitable visualization type based on the structure and semantics of the retrieved data. For example, temporal sequences are represented using line charts to emphasize trends over time, whereas categorical comparisons are rendered as bar or pie charts to highlight distributional contrasts. The Charting Expert generates executable visualization code dynamically, primarily using the Seaborn and Plotly libraries. The agent has access to a Python REPL environment, where the generated code is executed, and the resulting output or error stack trace is returned as structured feedback. This enables automatic error detection and self-correction—if a visualization fails due to syntax or data-related issues, the agent refactors the code or alters the chart type and re-executes it without human intervention. The combination of reasoning-guided chart selection, dynamic code synthesis, and feedback-driven refinement yields a visualization pipeline that is adaptive, self-correcting, and semantically aligned with both the data and the user’s intent.
\subsection{Explanation and Reasoning}

Beyond visualization, Polaris employs a dedicated Reasoning Expert to interpret query results and provide detailed insights. This agent leverages LLM to uncover latent patterns, identify causal relationships, and generate explanatory narratives, that go beyond descriptive summaries. Reasoning is further augmented with domain-specific knowledge extracted from the user’s database, allowing the system to produce explanations that are both contextually grounded and logically coherent \cite{wei2022chainofthought,wang2023selfconsistency}.

\subsection{Summarizing and Report Generation}

The Report Expert compiles insights, visualizations, and contextual information into structured reports. It aggregates content by summarizing query results, embedding visualizations, documenting methodologies, and incorporating metadata such as data sources and query parameters.

\section{Results}

We evaluate Polaris on the Airbnb listings dataset\cite{airbnbopendata_kaggle}, which contains structured metadata about rental properties across New York. To ensure systematic and reproducible evaluation, we employed a synthetic data generator to construct a golden dataset consisting of 40 query--answer pairs. Each instance comprises: (1) a natural language query, (2) a reference answer derived from the dataset, (3) the ground truth context used to generate the answer, (4) the retrieved context produced by our NL2SQL++ pipeline, and (5) the final agent response.

\subsection*{Evaluation Metrics}

\paragraph{Semantic Similarity.}
Semantic similarity measures the embedding-level similarity between retrieved context $c_r$ and ground truth context $c_g$, computed using cosine similarity:
\begin{equation}
\text{SemanticSimilarity}(c_r, c_g) = 
\frac{\langle \mathbf{e}(c_r), \mathbf{e}(c_g) \rangle}{\|\mathbf{e}(c_r)\| \, \|\mathbf{e}(c_g)\|}
\label{eq:semantic-similarity}
\end{equation}
where $\mathbf{e}(\cdot)$ denotes the embedding function.

\paragraph{Context Precision.}
Context precision quantifies the proportion of retrieved contexts that are relevant relative to the ground truth:
\begin{equation}
\text{ContextPrecision} = \frac{|C_r \cap C_g|}{|C_r|}
\label{eq:context-precision}
\end{equation}
where $C_r$ is the set of retrieved contexts and $C_g$ is the set of ground truth contexts.

\paragraph{Answer Relevancy.}
Answer relevancy evaluates whether the generated answer $a$ is relevant to both the query $q$ and the retrieved context $C_r$:
\begin{equation}
\begin{aligned}
\text{AnswerRelevancy}(a, q, C_r) 
&= \lambda\, \text{sim}(\mathbf{e}(a), \mathbf{e}(q)) \\
&\quad + (1-\lambda)\, \text{sim}(\mathbf{e}(a), \mathbf{e}(C_r))
\end{aligned}
\label{eq:answer-relevancy}
\end{equation}
where $\lambda \in [0,1]$ balances query alignment and context faithfulness.

\vspace{2pt}
\subsection{Interpretation}
The aggregated results across all 40 evaluation instances are summarized in Table~\ref{tab:avg_metrics}. Our system achieved high average scores across all metrics: semantic similarity (0.85), context precision (0.99), and answer relevancy (0.90).

\begin{table}[t]
    \centering
    \caption{Average metric scores across 40 evaluation samples.}
    \label{tab:avg_metrics}
    \begin{tabular}{lcc}
        \hline
        \textbf{Metric} & \textbf{Average Score} & \textbf{Model} \\
        \hline
        Semantic Similarity & 0.85 & GPT-4o \\
        Context Precision   & 0.99 & GPT-4o \\
        Answer Relevancy    & 0.90 & GPT-4o \\
        \hline
    \end{tabular}
\end{table}

To further analyze robustness, we evaluate the percentage of samples exceeding predefined metric thresholds (Table~\ref{tab:threshold_analysis}). Our framework demonstrated consistent precision and reliability, with 100\% of samples surpassing thresholds for semantic similarity and context precision, and 92.5\% surpassing the answer relevancy threshold.

\begin{table}[t]
    \centering
    \caption{Threshold-based evaluation of metric quality.}
    \label{tab:threshold_analysis}
    \begin{tabular}{lccc}
        \hline
        \textbf{Metric} & \textbf{Threshold} & \textbf{Above (\%)} \\
        \hline
        Semantic Similarity & 0.70 & 100.0 \\
        Context Precision   & 0.90 & 100.0 \\
        Answer Relevancy    & 0.70 & 92.5 \\
        \hline
    \end{tabular}
\end{table}

These results highlight two key findings. First, the NL2SQL++ module reliably captured user intent and generated precise SQL queries, as evidenced by perfect scores in semantic similarity and context precision. Second, while answer relevancy achieved a strong 92.5\% success rate, a small number of responses fell below the threshold. We attribute this variance to the generative nature of large language models, which may introduce stylistic or explanatory deviations even when grounded in correct evidence. Importantly, these deviations did not significantly degrade overall answer quality, underscoring the robustness of Polaris's multi-agent design for conversational enterprise analytics.

\section{Future Work and Conclusion}

We identify several directions to strengthen Polaris beyond the current implementation:
\begin{itemize}
    \item Adaptive orchestration with learned utility models and long term agent memeory that personalize $U(a,t \mid s)$ per organization and user.
    \item  Integrating with a global data catalog to ensure consistent and meaningful annotations across varied columns to maintain the quality of insights generated by AI agents.
    \item Workflows with editable plans, sandboxes for dry runs, and assisted what-if analyses.

\end{itemize}
    
We introduced Polaris, a supervisor-based multi-agent framework for conversational enterprise analytics centered on Dynamic Task Coordination. By formalizing orchestration as adaptive bipartite matching and coupling it with reason-first agents for querying, visualization, and explanation, the system executes coherent multi-step workflows with strong grounding. Our evaluation shows high semantic similarity and context precision, indicating faithful retrieval and reliable answer generation. These results suggest that principled orchestration is a promising path toward reliable, auditable, and efficient enterprise analytics at scale.

\bibliography{aaai2026}

@article{wang2023survey,
  title={A Survey on Large Language Model based Autonomous Agents},
  author={Wang, Lei and Ma, Chen and Feng, Xueyang and Zhang, Zeyu and Yang, Hao and Zhang, Jingsen and Chen, Zhiyuan and Tang, Jiakai and Chen, Xu and Lin, Yankai and others},
  journal={arXiv preprint arXiv:2308.11432},
  year={2023}
}

@INPROCEEDINGS{sql++,
  author={Carey, Michael and Chamberlin, Don and Goo, Almann and Ong, Kian Win and Papakonstantinou, Yannis and Suver, Chris and Vemulapalli, Sitaram and Westmann, Till},
  booktitle={2024 IEEE 40th International Conference on Data Engineering (ICDE)}, 
  title={SQL++: We Can Finally Relax!}, 
  year={2024},
  volume={},
  number={},
  pages={5501-5510},
  doi={10.1109/ICDE60146.2024.00438}}

@article{ReAct,
  title={{ReAct: Synergizing Reasoning and Acting in Language Models}},
  author={Yao, Shunyu and Zhao, Jeffrey and Yu, Dian and Du, Nan and Shafran, Izhak and Narasimhan, Karthik and Cao, Yuan},
  journal={arXiv e-prints},
  year={2023},
  eprint={2210.03629v3},
  note={Published in ICLR 2023}
}

@article{xi2023rise,
  title={The Rise and Potential of Large Language Model Based Agents: A Survey},
  author={Xi, Zhiheng and Chen, Wenxiang and Guo, Xin and He, Wang and Ding, Yiwen and Hong, Boyang and Zhang, Ming and Wang, Junzhe and Jin, Senjie and Zhou, Enyu and others},
  journal={arXiv preprint arXiv:2309.07864},
  year={2023}
}

@book{stone2000multiagent,
  title={Multiagent Systems: A Modern Approach to Distributed Artificial Intelligence},
  author={Stone, Peter and Veloso, Manuela},
  publisher={MIT Press},
  year={2000},
  address={Cambridge, MA}
}

@article{tampuu2017multiagent,
  title={Multiagent deep reinforcement learning with extremely sparse rewards},
  author={Tampuu, Ardi and Matiisen, Tambet and Kodelja, Dorian and Kuzovkin, Ilya and Korjus, Kristjan and Aru, J{\"u}ri and Aru, J{\"a}an and Vicente, Raul},
  journal={arXiv preprint arXiv:1707.01495},
  year={2017}
}

@inproceedings{wu2023autogen,
  title={AutoGen: Enabling Next-Gen LLM Applications via Multi-Agent Conversation},
  author={Wu, Qingyun and Bansal, Gagan and Zhang, Jieyu and Wu, Yiran and Zhang, Shaokun and Zhu, Erkang and Li, Beibin and Jiang, Li and Zhang, Xiaoxue and Wang, Chi},
  booktitle={Proceedings of the International Conference on Machine Learning},
  year={2023}
}

@inproceedings{hong2023metagpt,
  title={MetaGPT: Meta Programming for A Multi-Agent Collaborative Framework},
  author={Hong, Sirui and Zheng, Xiawu and Chen, Jonathan and Cheng, Yuheng and Zhang, Ceyao and Wang, Zili and Yau, Steven Ka Shing and Lin, Zijuan and Zhou, Liyang and Ran, Chenyu and others},
  booktitle={The Twelfth International Conference on Learning Representations},
  year={2023}
}

@misc{airbnbopendata_kaggle,
  title        = {Airbnb Open Data},
  author       = {Azmoudeh, Ariana},
  howpublished = {Kaggle},
  year         = {2021},
  note         = {Dataset available at \url{https://www.kaggle.com/datasets/arianazmoudeh/airbnbopendata}}
}

@article{shen2023hugginggpt,
  title   = {HuggingGPT: Solving AI Tasks with ChatGPT and its Friends in HuggingFace},
  author  = {Shen, Yongliang and Song, Kaitao and Tan, Xu and Jiang, Dongsheng and others},
  journal = {arXiv preprint arXiv:2303.17580},
  year    = {2023}
}

@article{schick2023toolformer,
  title   = {Toolformer: Language Models Can Teach Themselves to Use Tools},
  author  = {Schick, Timo and Dwivedi-Yu, Jane and Sch{"u}tze, Hinrich and others},
  journal = {arXiv preprint arXiv:2302.04761},
  year    = {2023}
}

@article{wei2022chainofthought,
  title   = {Chain-of-Thought Prompting Elicits Reasoning in Large Language Models},
  author  = {Wei, Jason and Wang, Xuezhi and Schuurmans, Dale and Bosma, Maarten and Ichter, Brian and others},
  journal = {arXiv preprint arXiv:2201.11903},
  year    = {2022}
}

@inproceedings{wang2023selfconsistency,
  title     = {Self-Consistency Improves Chain of Thought Reasoning in Language Models},
  author    = {Wang, Xuezhi and Wei, Jason and Schuurmans, Dale and Le, Quoc and Chi, Ed and Narang, Sharan and Chowdhery, Aakanksha and Zhou, Denny},
  booktitle = {International Conference on Learning Representations},
  year      = {2023}
}

@article{kuhn1955hungarian,
  title   = {The Hungarian Method for the Assignment Problem},
  author  = {Kuhn, Harold W},
  journal = {Naval Research Logistics Quarterly},
  volume  = {2},
  pages   = {83--97},
  year    = {1955}
}

@inproceedings{karp1990online,
  title     = {An Optimal Algorithm for On-line Bipartite Matching},
  author    = {Karp, Richard M and Vazirani, Umesh V and Vazirani, Vijay V},
  booktitle = {Proceedings of the Twenty-second Annual ACM Symposium on Theory of Computing},
  year      = {1990}
}


\end{document}